\documentclass[10pt]{cai}

\usepackage{amsmath,amssymb,amsfonts}
\usepackage{algorithmic}
\usepackage{acronym}
\usepackage{hyperref}
\usepackage{graphicx}
\usepackage{gensymb}
\usepackage{textcomp}
\usepackage{xcolor}
\usepackage{booktabs}

\begin{document}
\def\conferenceyear{2024}
\begin{center}

\title{GANsemble for Small and Imbalanced Data Sets:\\
A Baseline for Synthetic Microplastics Data}
\maketitle

\thispagestyle{empty}

\begin{tabular}{cc}
Daniel Platnick\upstairs{\affilone,\affiltwo,*}, Sourena Khanzadeh\upstairs{\affiltwo}, Alireza Sadeghian\upstairs{\affiltwo}, Richard Valenzano\upstairs{\affilone,\affiltwo}
\\[0.25ex]
{\small \upstairs{\affilone} Vector Institute, Toronto, Canada}, {\small \upstairs{\affiltwo} Toronto Metropolitan University, Toronto, Canada}\\
\end{tabular}

\emails{
  \upstairs{*}Corresponding author: daniel.platnick@torontomu.ca
}

\vspace*{0.2in}
\end{center}

\begin{abstract}

Microplastic particle ingestion or inhalation by humans is a problem of growing concern. Unfortunately, current research methods that use machine learning to understand their potential harms are obstructed by a lack of available data. Deep learning techniques in particular are challenged by such domains where only small or imbalanced data sets are available. Overcoming this challenge often involves oversampling underrepresented classes or augmenting the existing data to improve model performance. This paper proposes GANsemble: a two-module framework connecting data augmentation with conditional generative adversarial networks (cGANs) to generate class-conditioned synthetic data. First, the \emph{data chooser module} automates augmentation strategy selection by searching for the best data augmentation strategy. Next, the \emph{cGAN module} uses this strategy to train a cGAN for generating enhanced synthetic data. We experiment with the GANsemble framework on a small and imbalanced microplastics data set. A Microplastic-cGAN (MPcGAN) algorithm is introduced, and baselines for synthetic microplastics (SYMP) data are established in terms of Fréchet Inception Distance (FID) and Inception Scores (IS). We also provide a synthetic microplastics filter (SYMP-Filter) algorithm to increase the quality of generated SYMP. Additionally, we show the best amount of oversampling with augmentation to fix class imbalance in small microplastics data sets. To our knowledge, this study is the first application of generative AI to synthetically create microplastics data. 
\end{abstract}

\copyrightnotice

\section{Introduction}

The emergence of microscopic plastic particles in food and drinking water is a problem of growing concern \cite{mp_drinking_water}. Microplastics are shown as potentially harmful to reproductive health \cite{mp_reproductive}. They have been found to be increasingly present in human placenta \cite{mp_placenta} and human lungs  \cite{mp_lungs}. Thus, research towards mitigating potential dangers of microplastics is extremely valuable. However, such research can be difficult due to the insufficient availability of public data. This is particularly problematic for deep learning methods.  Despite this challenge, deep learning based methods have shown success for detecting microplastics even with small and imbalanced data sets \cite{source_dataset,data_source2}. To increase model robustness on small or imbalanced data sets, well known approaches such as creating augmented or synthetic data can be used to oversample minority classes or increase data set size. We connect these ideas with GANsemble, which performs a search for the best augmentation strategy, and uses it to train a cGAN and create class-conditioned SYMP data.

Deep learning techniques have proven value in many domains \cite{deep_theorem_prover,nlp,DINR}, but universally share weakness in the cases of small data sets and those with class imbalances \cite{small_data,class_imbalance}. The data a deep learning model is trained on greatly determines it's robustness, scalability, safety, and fairness \cite{fairness}. Data augmentation and transfer learning have been used to increase the effectiveness of deep learning techniques in the cases of small or imbalanced training data sets \cite{augmentation_survey}. However, data augmentation only works to a certain extent, and transfer learning can have the side effect of negative transfer \cite{negative_transfer}. As such, using generative AI to create synthetic training data is an increasingly popular strategy \cite{GAN-ST}. Generative AI approaches for creating synthetic data are shown to be useful across different domains, as they can generate an arbitrary number of synthetic samples, and the synthetic data is not affected by privacy regulations \cite{privacy_reg}. As generative AI research continues to progress, the quality of generated synthetic samples continues to improve, which increases their effectiveness for training deep learning models.

Oversampling the minority class is a natural remedy for class imbalance \cite{Oversample}. Synthetic Minority Oversampling Technique (SMOTE) is a well known method which produces synthetic samples to fix class imbalances using K-nearest neighbors in the feature space \cite{SMOTE}. \citeauthor{heart_spectral_aug} \cite{heart_spectral_aug} performed analysis on different augmentation strategies for convolutional neural network (CNN) classification of medical image spectra, and found that simple strategies such as masking and horizontal flipping can greatly improve model performance. Today, oversampling techniques typically include intelligent data augmentations \cite{augmentation_survey}. \citeauthor{CutMix} \cite{CutMix} introduced a state-of-the-art image augmentation algorithm called CutMix, which maintains the benefits of regional dropout regularization without loss of information. \citeauthor{no_best_resampling_approach} \cite{no_best_resampling_approach} showed that cases may vary depending on the nature of the data. Therefore, given imperfect information, the system designer must decide the best augmentation strategy to use. This is the challenge of augmentation selection.

Inspired by recent developments in generative AI \cite{GAN,cGAN}, we present GANsemble to overcome this challenge and further microplastics research. The proposed method connects ideas from data augmentation and synthetic data generation \cite{cGAN_synthdata} to provide automatic augmentation strategy selection and enhance model learning on small and imbalanced data sets. The GANsemble \emph{data chooser module} enables the AI system to manipulate the inputted data through augmentation, generating new augmentation strategies to find the strategy which maximizes model generalization. Specifically, GANsemble performs an \emph{n-step factorial search} search on inputted augmentation strategies, and employs a deep neural network (DNN) to intelligently select the best strategy $Aug^*$. This strategy is then used to oversample the original data set and train a cGAN to generate new synthetic samples. GANsemble can significantly improve the augmentation selection process, by decreasing the need for human intervention. Furthermore, we provide MPcGAN, the first cGAN for generating synthetic microplastics (SYMP) data, and establish baseline FID and IS scores for SYMP data.

Generating synthetic sample data to enhance the training set is an idea increasing in popularity \cite{cGAN,diffusion_sample,diffusion_sample,CVAE}. This approach has not been applied to microplastics research. To make progress in SYMP data generation, this work applies cGANs to create SYMP data, and baseline SYMP data quality measures are established. MPcGAN is trained to produce class-conditioned SYMP samples that enhance training set size, balance, and variation. Note, this paper builds on the important work of \citeauthor{source_dataset} \cite{source_dataset}, which identified the most effective amount of oversampling with augmentation for a small imbalanced microplastics data set. However, Tian et al. introduce data leakage into their analysis by creating augmented data from the training set, and evaluating on the same training set. This is a problem, because in practice, the model will be evaluated on data it has never seen before, not data created from previous training data. Performing a similar study, this research builds on their analysis by using a separate evaluation set to prevent data leakage between train and test sets. Multiple runs are used to increase experimental validity.

The rest of this paper is organized as follows. The microplastics data description is given in Section \ref{sec:datadesc}. Section \ref{sec:methodology} explains the GANsemble framework, as well as techniques, algorithms, and parameters used in the study. Section \ref{sec:experiments_results} provides experiment results, and establishes baseline FID and IS scores for SYMP data. Next, section \ref{sec:discussion} discusses the implications as well as concluding remarks.

\section{Data Set Description}
\label{sec:datadesc}

The data set we use was introduced by \citeauthor{source_dataset} \cite{source_dataset}, and is the only easily accessible microplastics data set we could find.
The raw data and supplementary information on data preparation can be found from these sources \cite{source_dataset,data_source2,data_source3}. 
In this section, we review its characteristics and how it was created. 

 \begin{figure*}[t]
     \centering
     \includegraphics[scale=0.43]{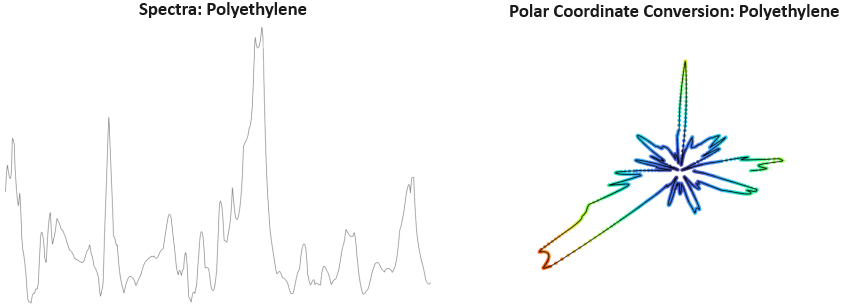}
     \caption{Example of the polar conversion that allows DNNs to treat polymer spectra as images \cite{source_dataset}.}
     \label{PLATNICK.fig01.png}
 \end{figure*}
 
Laser Directed Infrared (LDIR) and Fourier Transform Infrared (FTIR) spectroscopy convert chemical compounds to spectra data, which has previously been shown amenable for learning by deep neural networks (DNNs) \cite{CNN_soil,CNN_spectra,LDIR}.  The data set used in this study contains 210 samples of polymer spectra across 10 classes: 8 classes of plastic spectra, and 2 non-plastic classes \cite{source_dataset}. The least represented polymer class, polyacetal, has 5 samples, while the most highly represented class silica has 38. The 210 particle-based spectra were collected from various Dutch water sources such as drinking water, surface water, and sewage, as well as the Agilent Clarity version 1.4.10 software database \cite{data_source2}. Samples were converted to spectra using the LDIR quantum cascade laser (QCL)-based Agilent chemical imaging system. The LDIR system was used with wavenumbers between 975 and 1800cm$^{-1}$, and a rotation of $0.5$cm$^{-1}$ \cite{data_source3,source_dataset}.
An example spectra can be found in the left image of Figure \ref{PLATNICK.fig01.png}.

Tian et al. \cite{source_dataset} also introduced a visual polarized coordinate transformation which allows microplastic particle spectra to be treated as as images. This enables the use of computer vision models, which have been found to be very useful in practice.
The spectra in the microplastics data set are mapped through the polar conversion process as follows: The wavenumbers 975-1800 are mapped from 0$\degree$ to 360$\degree$ and the absorption rate becomes the polar coordinate magnitude. The images are then colourized such that colour intensity grows with the absorption value. Figure \ref{PLATNICK.fig01.png} shows an example of this transformation visually.

\section{The GANsemble Framework}
\label{sec:methodology}

In this section, we describe our GANsemble framework and how it is applied to the microplastics data set.

GANsemble consists of two distinct detachable modules, which we describe below.
The first, the \emph{data chooser module}, is used to identify useful combinations of base augmentation strategies, which we call \emph{composite strategies}, for the given data set.
Next, we have the \emph{cGan module}, which uses the best base augmentation strategy or \emph{composite strategy} to generate more data for the training of a conditional GAN.
Finally, we have the \emph{SYMP-Filter}, which post-processes the output of the cGAN by removing artifacts of generated images, thereby improving the quality of the synthetic data.
We describe these modules below.

\subsection{The Data Chooser Module} \label{sec:data_chooser} 
The first component of the GANsemble framework is the \emph{data chooser module}. The objective of this module is to automate augmentation selection by searching for the augmentation strategy or \emph{composite strategy}, given a set of base augmentation strategies, that can be used to most effectively train a cGAN on a small and possibly imbalanced data set.
Since this process involves finding effective combinations of base augmentation strategies, which we call \emph{composite strategies}, the module has several hyperparameters controlling the amount of computational resources to be used in the search for the best augmentation strategy, $Aug^*$.
Thus, the \emph{data chooser module} simplifies the challenge of finding the best augmentation strategy from  ``What augmentation strategy should I use?" to ``How many augmentation strategies should I consider and how extensively should they be considered, given time and computational resources?"

In more detail, the \emph{data chooser module} takes in 3 inputs. The first is a small imbalanced data set requiring augmentation. In our case, this is the microplastic spectra images. 

The second input is a list of base augmentation strategies to be used. While any set of base data augmentation strategies can be used, for the microplastic data, we use four Gaussian approaches: (1) horizontal flipping and horizontal shifting, (2) blur and rotation, (3) zoom and rotation, and (4) circular and rectangular masking. Base augmentation strategy 1 performs random horizontal flipping and shifting. Random horizontal flips in spectra data have been shown to greatly improve ML classification performance \cite{heart_spectral_aug}, while shifting ensures there are not too many duplicate samples. Augmentation strategy 2 uses varying levels of blur and random rotations to distort the spectra images. The blur effect is achieved through a Gaussian filter with $\alpha \in [10,23]$ and $\sigma \in [2.8,3.82]$, where $\alpha$ is blur intensity and $\sigma$ is smoothness. Different $\alpha$ and $\sigma$ values were tried and these ranges gave the best results. Augmentation strategy 3 applies random rotations and zoom between $1.0$x and $1.34$x. Augmentation strategy 4 is a Monte Carlo style masking strategy, where circles and rectangles cover different areas of the polymer spectra. Masking has also been effective for augmenting data in heart spectra classification \cite{heart_spectral_aug}, as it forces the learning algorithm to focus on different regions of the image when masked regions are not present.

The third input to the \emph{data chooser module} is $n$, the number of steps to take in the \emph{n-step factorial search}. Given a data set and a set of base augmentation strategies, the \emph{data chooser module} identifies which base augmentation strategy or \emph{composite strategy} provides the most useful data set. However, there are $\mathcal{O}(2^A)$ total strategies, given $A$ base augmentation strategies.
Considering all of them may be prohibitively expensive, which is why $n$ is introduced.
For a given $n$, only combinations containing at most $n$ augmentation methods are considered.
This decreases the number of combinations to $\mathcal{O}(2^n)$.

In principle, it would be possible to train all $\mathcal{O}(2^A)$ possible cGANs and thereby directly evaluate the effectiveness of the different combinations.
However, even for a small data set, doing so is prohibitively expensive given the time and lack of stability of GAN training.
As such, we instead focus on the simpler task of classification using a pre-trained model.
That is, we consider each of the $\mathcal{O}(2^n)$ possible combinations of augmentation methods, and train a classifier $r$ times each, where $r$ is another hyperparameter controlling how extensively the augmentation strategies are explored.
The classifier with the best average performance over the $r$ runs is then chosen and used for training the cGAN. For the microplastics data, we use a pre-trained ResNet50 as the classifier for identifying the best augmentation strategy.
We refer to the resulting augmentation strategy as $Aug^*$.



\subsection{The cGAN Module}

The next module is the cGAN trained on $Aug^*$. We propose a baseline MPcGAN model for microplastics spectra data to showcase GANsemble and further generative AI research in mitigating microplastics. The \emph{cGAN module} attaches to the \emph{data chooser module}, training a generator on the \emph{data chooser module} output to produce robust synthetic training data. As input, the MPcGAN generator takes in class label $y$, and samples from the Gaussian distribution $z$ conditionally based on $y$, to produce a synthetic image from class $y$. The discriminator takes in a class label with a real or synthetic image and outputs a value $D(G(z|y))\in [0,1]$, where values closer to 1 correspond to the discriminator predicting the data is real. Both the generator and discriminator architectures are based on the convolutional downsample and upsample operations. Information on the cGAN training setup can be found in section \ref{sec:experiments_results}.

\subsection{The SYMP-Filter}
Finally, we provide a post-processing algorithm called the \emph{SYMP-Filter}. This process results in higher quality synthetic data. Our algorithm leverages spatial attributes of the polar coordinate transformed images to detect and filter out noisy SYMP data points. Figure \ref{fig:Filter} illustrates the effects of the \emph{SYMP-Filter} visually.

The algorithm works through this process: Generate 5000 SYMP from each class, rank them using the \emph{SYMP-Filter}, and keep the top $t$ images of each class. The \emph{SYMP-Filter} applies a square mask to the 4 corner sections of each image, calculating the densities of each corner. The pixel values inside the corner sections are summed to get the corner density for each image. Next, the algorithm keeps the top $t$ lowest density images from each class.

The \emph{SYMP-Filter} algorithm is described mathematically as follows:
\begin{enumerate}
    \item Let $X_{i,j}$ denote an image in a collection, with $i$ indexing the collection and $j$ indexing the images within that collection, where each image is a 3-dimensional tensor in $\mathbb{R}^{128 \times 128 \times 3}$.
    \item Let $r$ be the radius defining the area of the corner sections to calculate each density.
    \item Let $M_k(r)$ represent the square mask density detector for the $k$-th corner section, with $k \in \{1, 2, 3, 4\}$ corresponding to the four corner sections and radius $r$.
    \item Let $D_{i,j}$ denote the density for image $X_{i,j}$, computed by applying the corner masks and summing the values.
    \item Let $T$ be the set of indices $(i, j)$ corresponding to the top $t$ images with the lowest densities.
\end{enumerate}
\begin{equation}
\label{density_formula}
D_{i,j} = \sum_{k=1}^{4} \sum_{x=1}^{128} \sum_{y=1}^{128} \sum_{c=1}^{3} X_{i,j}(x, y, c) \cdot M_k(x, y, c, r)    
\end{equation}

\begin{equation}
    \label{Top_den_formula}
    T = \underset{i,j}{\mathrm{argmin}_t} \left( D_{i,j} \right) 
\end{equation}



\section{Experiments and Results}
\label{sec:experiments_results}


Recall that our data set contains only 210 images. This was split into an unbiased testing set containing 20 samples, with 2 images per class, and a training set containing the remaining 190 samples. The 20 samples were not used in creating the augmented data sets from Table \ref{tbl:dcm_table} or Figure \ref{PLATNICK.fig02.png} to prevent data leakage. The 20 samples were later added to help train the cGAN. All code relating to this work can be found here: 
\url{https://github.com/DanielPlatnick/GANsemble}.

\subsection{Identifying An Effective Size For Augmented Data Sets}

 \begin{figure*}[t]
     \centering
     \includegraphics[scale=0.45]{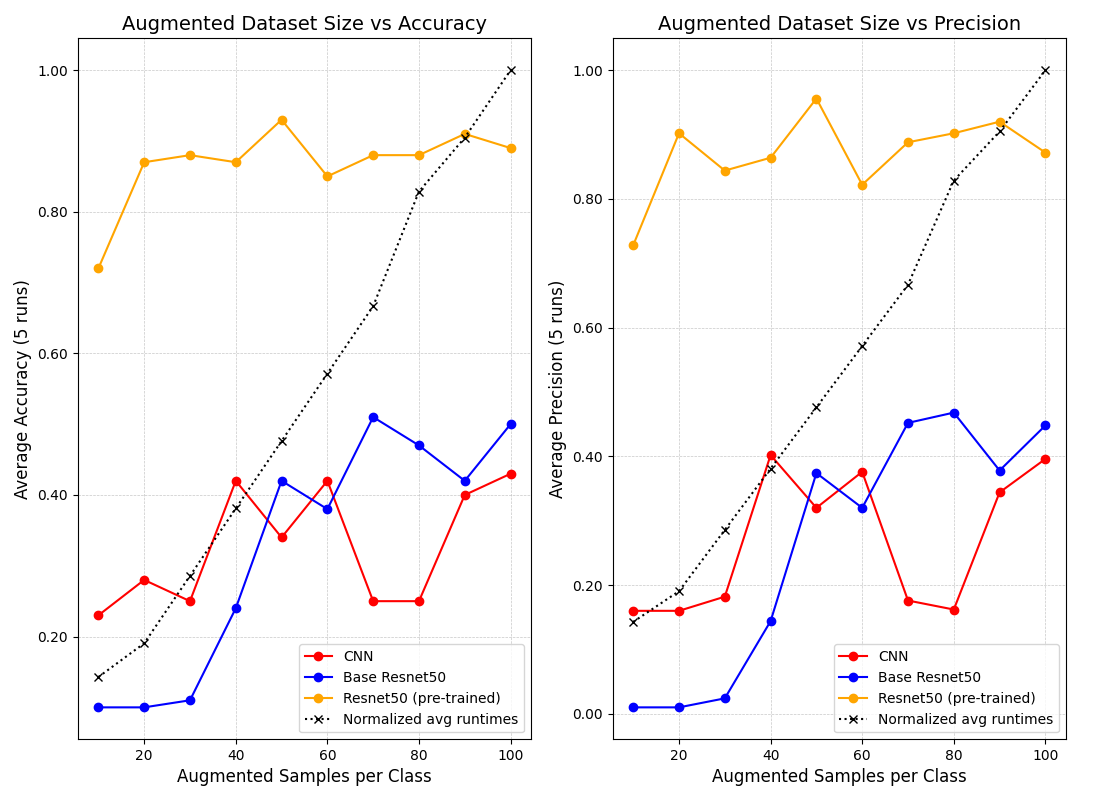}
     \caption{Determining the most effective size for purely augmented small microplastics data sets. The black dotted line represents averaged algorithm run-time after min-max normalization. The best performance is achieved with a pre-trained ResNet50 model using 50 augmented samples from each class. Augmentation strategy 1 was used to generate figure \ref{PLATNICK.fig02.png}.}
     \label{PLATNICK.fig02.png}
 \end{figure*}
 
In our first experiment, we attempt to reproduce the experiments by Tian et al. \cite{source_dataset}, in which they evaluated the tradeoff between accuracy and training time related to augmented data set size. In the original work, a custom augmentation strategy was used. Our goal is to evaluate if similar conclusions can be made with simpler augmentation strategies and without the data leakage that occurred in the original work.

For simplicity, we tested with horizontal flipping and horizontal shifting. Using this augmentation strategy, we create 10 purely augmented data sets of sizes $100, 200, \ldots, 1000$, each with balanced classes.
We then used this data to train three different DNN models: a CNN with $13,804,510$ trainable parameters, a ResNet50 with $23,528,522$ trainable parameters, and a version of that ResNet50 pre-trained on ImageNet \cite{Resnet, ImageNet}. 
Each DNN was trained 5 times and the average results on our test set of 20 images are shown in Figure \ref{PLATNICK.fig02.png}.
The average training time over all models is also shown. Training times are normalized between 0 and 1, where 0 corresponds to the lowest runtime of any model, and 1 corresponds to the largest runtime.

Figure \ref{PLATNICK.fig02.png} shows that the pre-trained ResNet50 is most capable of generalization based on accuracy and precision. Transfer learning is a well-known method for enhancing model generalization, and should be used when applicable \cite{negative_transfer}. The pre-trained ResNet50 achieves the highest performance with $50$ samples per class (\textit{i.e.} 500 total augmented samples), scoring an average accuracy of 93\% and precision of 95\%. Run-time increases linearly with data set size, and the best size in terms of performance and run-time is $n=50$. Even with 10 samples, the pre-trained ResNet is capable of learning enough to achieve 75\% accuracy. The other models had a much higher variance across runs, and greatly lagged behind in performance at the levels of data tested.

Interestingly, our results roughly correspond with those originally by \citeauthor{source_dataset} \cite{source_dataset}.
In their work, the best performance was found when using $40$ samples per class.

\subsection{Automating Augmentation Selection}

Next, we use the \emph{data chooser module} to automate augmentation strategy selection by identifying the best augmentation strategy or \emph{composite strategy} to be used in cGAN training.
The 4 base augmentation strategies used were described in Section \ref{sec:data_chooser}.
We set the number of steps $n=4$ and thus considered $15$ different augmentation methods in total.
These 15 strategies were used to create 15 data sets containing $50$ images per class, with a mixture of augmented and real samples.
For example, if $10$ real samples were available for a particular class, then $40$ augmented samples were added to that class.

\begin{table}[t]
    \centering
    \begin{tabular}{r|r}
        \toprule
        \multicolumn{1}{c|}{\textbf{Data Set}} & \multicolumn{1}{c}{\textbf{Accuracy ($\%$)}} \\
        \midrule
        \textbf{Aug 1} & \textbf{90.5} \\
        Aug 2 & 79.5 \\ 
        Aug 3 & 84.0 \\ 
        \textbf{Aug 4*} & \textbf{91.5} \\
        Aug 5 (1 \& 2) & 85.0  \\ 
        Aug 6 (1 \& 3) & 85.5 \\ 
        \textbf{Aug 7 (1 \& 4)} & \textbf{89.5}  \\ 
        Aug 8 (2 \& 3) & 88.5  \\ 
        Aug 9 (2 \& 4) & 87.0  \\ 
        Aug 10 (3 \& 4) & 85.0 \\ 
        Aug 11 (1, 2, \& 3)& 85.0  \\ 
        Aug 12 (1, 2, \& 4)& 87.0 \\ 
        Aug 13 (1, 3, \& 4)& 84.5  \\ 
        Aug 14 (2, 3, \& 4)& 88.5 \\ 
        Aug 15 (1, 2, 3, \& 4)& 84.0 \\ 
        \midrule
        No Oversampling & 86.0 \\ 
        Oversampling No Aug & 87.5\\
        \bottomrule
    \end{tabular}
    \caption{Demonstration of the GANsemble \emph{data chooser module} \emph{n-step factorial search}. Based on Figure \ref{PLATNICK.fig02.png} experiments, $N = 500$ was used to oversample the data. Performances are measured in terms of average accuracy over 10 runs. The three top \emph{data chooser module} (pre-trained ResNet50) choices are highlighted for analysis. GANsemble detects the best augmentation strategy $Aug^*$. \emph{Oversampling No Aug} uses duplication.}
\label{tbl:dcm_table}
\end{table}

For each of the $15$ resulting data sets, we fine-tuned a pre-trained ResNet50 $10$ times for 150 epochs, training on the $190$ samples and evaluating on the $20$ test samples. The results are shown in Table \ref{tbl:dcm_table}, along with the performances of the ResNet50s when trained using oversampling on the original data to fix class imbalance (\emph{Oversampling No Aug}), and when using the original data directly (\emph{No Oversampling}). The top three augmentation strategies are 1, 4 ($Aug^*$), and 7. Strategy 1 applies horizontal flipping and shifting, $Aug^*$ uses masking, and strategy 7 is a \emph{composite strategy} of strategies 1 and $Aug^*$.

 \begin{figure*}[htb]
     \centering
     \includegraphics[width =\textwidth]{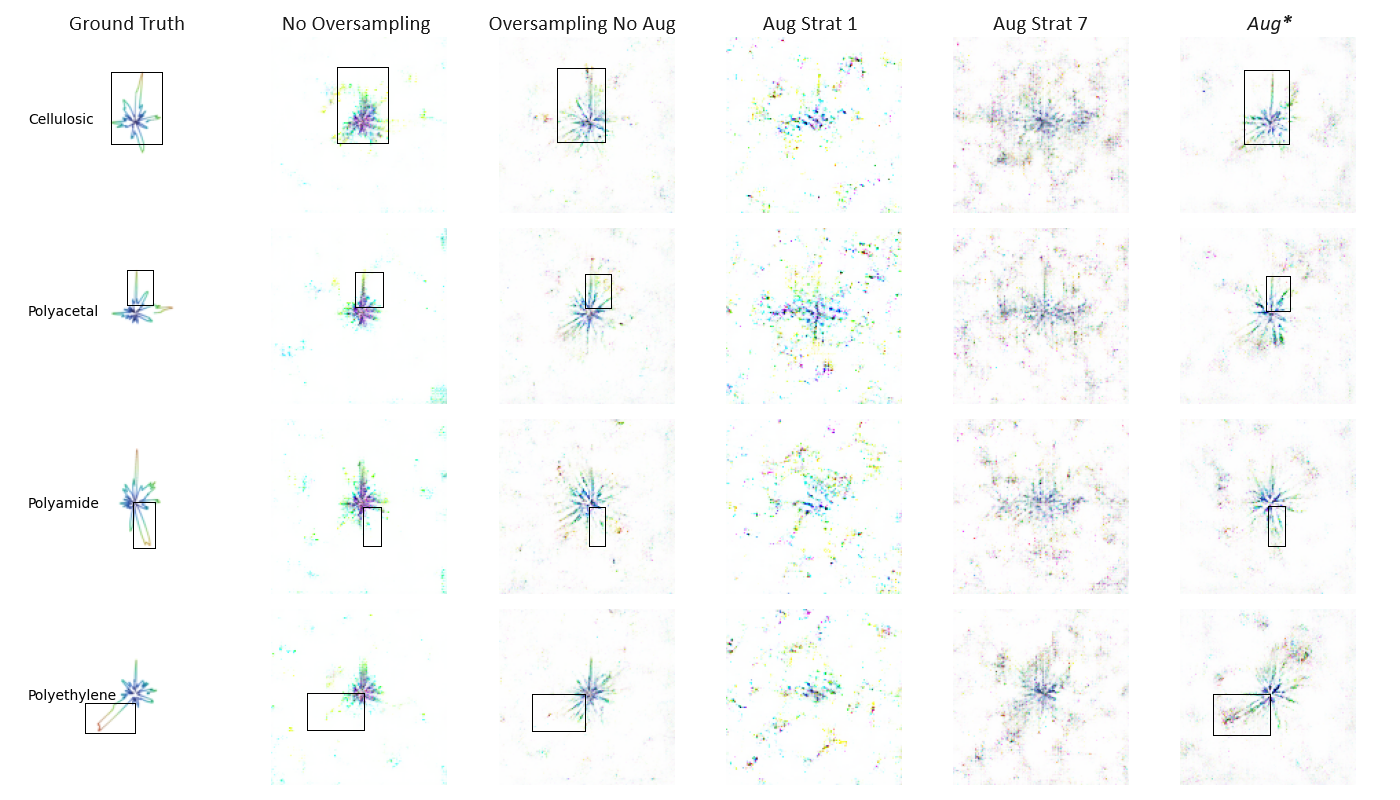}
     \caption{Qualitative results of SYMP data generated from MPcGAN instances trained on the top 3 strategies and baselines. We include SYMP samples from the first 4 classes. Boxes are drawn to highlight significant class-dependent features. The 20 samples were generated in a batch and not cherry picked. MPcGAN trained on data oversampled with $Aug^*$ results in the best SYMP in terms of ground truth resemblance.}
     \label{PLATNICK.fig03.png}
 \end{figure*}

Random masking results in the highest accuracy. Many \emph{composite strategies} improve classification performance, but some hurt model performance. We find that augmentation strategies affecting the spatial domain in different ways should not be combined. Other notable \emph{composite strategies} include strategies 8, 9, 12 and 14. Augmentation strategy 14 performs well and is a combination of strategies 2, 3, and 4. This implies certain augmentation strategies complement each other, and GANsemble is capable of detecting these complements. Our automation strategy is general and not limited to microplastics data.


\subsection{cGAN Performance}

After $Aug^*$ is identified, our system uses it to create augmented data to train a cGAN, which we call the microplastics cGAN or MPcGAN.
In this section, we consider the performance of MPcGAN.

For this test, we trained 5 different MPcGAN models, each on a different data set.
These 5 data sets were crafted using the top 3 performing augmentation strategies (strategies 1, 4, and 7), balanced oversampling on the $210$ total real images (\emph{Oversampling No Aug}), and the unbalanced data set consisting only of the real images (\emph{No Oversampling}). The augmented data sets consisted of the $210$ real images and $390$ augmented images to create $500$ total images balanced across classes.

For each cGAN model, during training iterations, the discriminator is trained on a batch of size 128, consisting of 64 real samples and 64 synthetic samples generated by the generator. The generator is trained by passing generated images through the frozen discriminator, and using the discriminator output as an error signal. A train$-$test split of 0.86$-$0.14 was used to train the discriminator. MPcGAN is designed to generate images of size 128x128x3. The best found settings for the generator and discriminator learning rates were 0.002, and 0.0002, respectively. The generator and discriminator both use Adam optimizer with binary cross entropy as the loss function \cite{Adam}. Dropout is used in the discriminator for regularization, and Leaky ReLus are used for activation functions in the cGAN \cite{Dropout, Leaky}.

MPcGAN collapses often, and many attempts were made to see how many epochs each training approach could endure before collapsing. We found that MPcGAN can train for roughly 150 epochs before collapsing when using the strategies $Aug^*$, \emph{Oversampling No Aug}, or \emph{No Oversampling}. While training MPcGAN on strategies 1 or 7, the cGAN is unstable and can only train up to 80 epochs before collapsing. Augmentation strategy 1 applies horizontal flipping and shifting. \emph{Composite strategy} 7 uses horizontal flipping and shifting combined with random masking. The unstable training is likely due to the effects strategies 1 and 7 have on the image spatial dimension, which cause the cGAN to lose focus during training, collapsing the network. Small data cGAN training is unstable, and MPcGAN requires multiple attempts at convergence, converging about 16\% of the time.

Figure \ref{PLATNICK.fig03.png} shows qualitative results of synthetic microplastic spectra images from 5 trained MPcGAN instances, each trained with different approaches. Synthetic spectra of 4 classes were generated: Cellulosic, polyacetal, polyamide, and polyethylene. Bounding boxes are drawn over class-dependent visual features of each polymer class. Overall, we found that MPcGAN trained on $Aug^*$ outperforms other methods in terms of generated SYMP feature variance, visual quality, complexity, and ground truth resemblance.

In the bounding box of the ground truth polyethylene sample in figure \ref{PLATNICK.fig03.png}, polyethylene has a characteristic left-downward diagonal peak with a red tip. This feature is common in the real images, and robust synthetic samples should inherit this feature. MPcGAN trained on $Aug^*$ is the only method capable of producing this feature. This feature learning capability is also present in other examples. For example, polyamide (row 3) has a characteristic bottom-right diagonal peak, which $Aug^*$ best mimics.

We also notice that $Aug^*$ based SYMP better follows ground truth colouring by adding whiteness near the center, while baselines tend to fill the center with blue. Synthetic spectra generated from $Aug^*$ are superior in quality with higher feature variance. SYMP generated without augmentation or oversampling lack variation. Samples generated with oversampling but no augmentation gain slight variation, but do not learn class specific features like samples generated from ${Aug^*}$.

\begin{figure*}[t]
     \centering
     \includegraphics[scale=0.25]{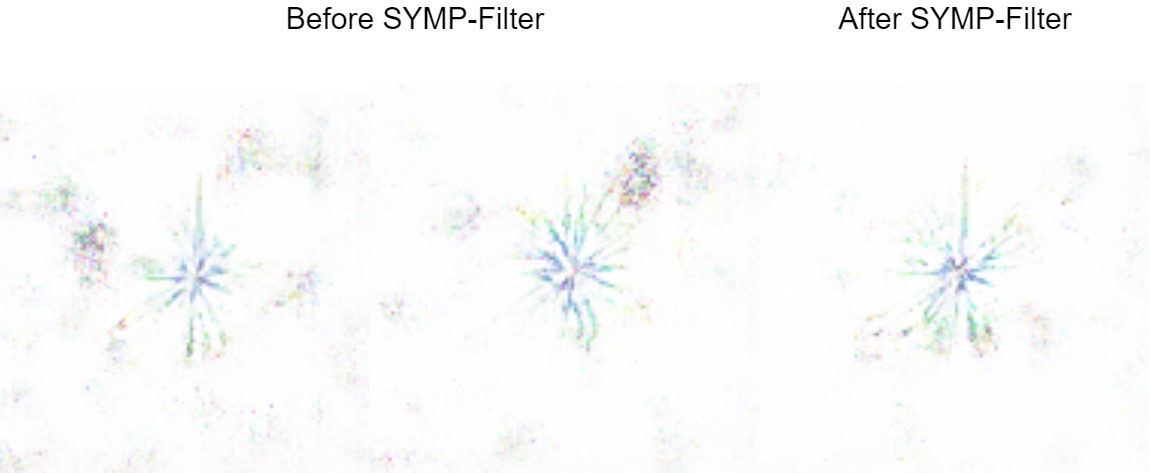}
     \caption{Qualitative results of our \emph{SYMP-Filter} algorithm. Three synthetic images of Cellulosic were generated using a cGAN trained on $Aug^*$. The two images on the left contain unprocessed cGAN output, while the rightmost image shows filtered output.}
     \label{fig:Filter}
 \end{figure*}

\begin{table}[b]
    \centering
    \begin{tabular}{r|rr|rr|rr}
    \toprule
    &\multicolumn{2}{c|}{\textbf{FID}}&\multicolumn{2}{c}{\textbf{IS Mean}}&\multicolumn{2}{c}{\textbf{IS Stdev}}\\
    \textbf{Data Set}&\textbf{Regular}&\textbf{Filtered}&\textbf{Regular}&\textbf{Filtered}&\textbf{Regular}&\textbf{Filtered}\\
    \midrule
         Aug*               & 272.9 & 247.9 & 1.34 & 1.22 & 0.026  & 0.030 \\
         Aug 1              & 288.2 & 287.1 & 1.13 & 1.11 & 0.012  & 0.018 \\
         Aug 7              & 395.5 & 392.0 & 1.20 & 1.21 & 0.037  & 0.021 \\
         No Oversampling    & 216.0 & 214.3 & 1.14 & 1.12 & 0.062  & 0.011 \\
         Oversampling No Aug& 261.0 & 226.5 & 1.31 & 1.16 & 0.028  & 0.018 \\
    \end{tabular}
    \caption{Establishing baseline FID and Inception Scores (IS) for SYMP data generation. Parentheses describe performances after applying the SYMP-Filter algorithm. Five MPcGAN training approaches are examined. Low FID and high IS are desirable.}
\label{tbl:FID_IS_MP_baselines}
\end{table}


We have computed Fréchet Inception Distance (FID) and Inception Scores (IS) scores for SYMP data in Table \ref{tbl:FID_IS_MP_baselines}. FID and IS are two metrics used to evaluate the quality of synthetic data based on the available ground truth data distribution \cite{FID_IS}. Low FID and high IS correspond to better sample quality. Low FID means the synthetic data better resembles the ground truth. High IS ensures that samples have high variance and clarity. The table also shows the performance before (\emph{Regular}) and after (\emph{Filtered}) applying the SYMP-Filter.

Table \ref{tbl:FID_IS_MP_baselines} shows $Aug^*$ achieves the best balance between FID and IS scores.
The best FID is seen with no oversampling or augmentation, but this is due to the mostly white backgrounds in the SYMP images generated by these methods, meaning MPcGAN is unable to generate samples with variance, resulting in lower ISs. 
Oversampling without augmentation generally has strong results, but a significantly weaker IS than $Aug^*$ in terms of both mean and standard deviation. No oversampling and oversampling without augmentation both resulted in strong empirical performance, but significantly lack variation. MPcGAN trained on the \emph{data chooser module} top choice data set $Aug^*$ produces the best SYMP samples when considered qualitatively, and also has a desirable balance of FID and IS performance.


\vspace{-0.25cm}
\section{Discussion and Conclusions}
\label{sec:discussion}
Microplastics are a problem of growing concern, and current research on the problem is obstructed by limited available data. Our work on GANsemble addresses the issues of small data and class imbalance by automating data augmentation selection and improving synthetic sample generation. We show the \emph{data chooser module} can be used to automate augmentation selection on a small microplastics data set, and believe it should work in other small data settings as well. MPcGAN is a fairly simple generator and discriminator architecture, and enhancements to the architecture would improve the SYMP data generated. Future work should include implementations of GANsemble with augmentation strategies that do not apply spatial transformations to the image. New augmentations for GANsemble could also include variations of blurring, random erasing \cite{random_erase}, or the CutMix algorithm \cite{CutMix}.

We demonstrate the ability of GANsemble to automate augmentation selection and establish baselines for SYMP data generation in terms of FID and IS scores. The GANsemble \emph{data chooser module} is capable of identifying an augmentation strategy which qualitatively and quantitatively outperforms all other compared methods. This is useful in domains where data collection is costly or there is low economic incentive. Our proposed MPcGAN algorithm is capable of learning to generate robust SYMP samples inheriting class dependent features. We encourage others to build on this work to further efforts in microplastics identification and mitigation. 

\section*{Acknowledgements}

We thank the anonymous reviewers for their feedback on this work. We also gratefully acknowledge the financial support of the Natural Sciences and Engineering Research Council of Canada (NSERC).

\printbibliography[heading=subbibintoc]

\end{document}